# Unlocking UML Class Diagram Understanding in Vision Language Models

Artem Naboichenko[1] and René Peinl[1]

[1] Hof University of Applied Sciences, Hof, Germany
artem.naboichenko | rene.peinl@hof-university.de

**Abstract.** Although Vision Language Models (VLMs) have seen tremendous progress across all kinds of use cases, they still fall behind in answering questions regarding diagrams compared to photos. Although progress has been made in the area of bar charts, line charts and other diagrams like that there is still few research concerned with other types of diagrams, e.g. in the computer science domain. Our work presents a benchmark for visual question answering based on UML class diagrams which is both challenging and manageable. We further construct a large-scale training dataset with 16.000 image-question-answer triples and show that a LoRA-based finetune easily outperforms Qwen 3.5 27B, which is a recent and well-performing VLM in many other benchmarks.

## 1 Introduction (Heading 1)

Vision Language Models (VLMs), i.e. the combination of a large language model (LLM) with a vision encoder using a projection layer to bring visual understanding to text-based models, have made tremendous progress in comprehension of photos. The typically used image encoders (often CLIP [1] or SigLIP [2], seldom others like DINO [3], [4]) are pre-trained on real world images and therefore excel on their understanding, especially for scene description and identifying objects and their relation on the image. However, despite a couple of specialized benchmarks like ChartQA [5], VLMs still struggle with interpreting diagrams as soon as it goes beyond questions that can be mainly answered with OCR (optical character recognition) capabilities [6].

Although coding is a major topic for LLMs, the understanding of code related diagrams like UML [7] or BPMN [8] is still underexplored. Astonishingly, the generation of UML diagrams using LLMs has attracted much more attention in scientific literature [9], [10], [11], [12], [13], [14], [15].

In this paper we explore the shortcomings of current VLMs on answering questions related to UML class diagrams and show that these issues are not of principle nature but can be mitigated with LoRA (Low Rank Adaptation, [16]) with sufficient training data. We present a large dataset collected from real world software repositories with Java program code that we use to finetune Qwen 2.5 VL 7B using various methods.

Our best finetune outperforms the recently published Qwen 3.5 27B by 20%, reaching 85.9% accuracy. The results of the Qwen 2.5 VL 7B baseline (44%) and Qwen 3 VL 8B (54%) show that the dataset is challenging, but scores improve well in line with general visual capabilities as measured by, e.g. MMMU Pro (Qwen 2.5 VL 7B: 41%, Qwen 3 VL 8B: 55.9%, Qwen3.5 27B: 75%).

The paper is structured as follows. First, related work is analyzed and gaps in the current research are identified. After that, the dataset is introduced and the setup of the conducted experiments are described. After that, results are presented, comparisons between finetunes and available open weight models are made. Mistakes of the VLMs are categorized and analyzed in quite some detail. We conclude with a discussion and future work.

# 2 Related Work

To the best of the authors' knowledge there is no prior work that deals with question answering on images showing UML class diagrams. However, there are both related works dealing with generating UML class diagrams from text [9], [12], [13] as well as interpretation of other types of diagrams like BPMN [8] or charts [6], [17] by VLMs. The closest related work we could find is [7].

## 2.1 UML generation with LLMs and VLMs

UML class diagrams are among the most widely used artifacts in object-oriented modeling [9]. They statically represent the internal structure of a software system by identifying its main components (classes), the data they contain (attributes), the operations that can be performed on the data (methods), and the relationships between them (associations, aggregations, compositions, and generalizations)[9]. Babaalla et al. use small LLMs like BERT and Electra to extract relevant keywords from a natural language specification via Named Entity Recognition (NER) as part of a model-driven architecture (MDA) initiative. They algorithmically transform the named entities into a textual domain-specific language (DSL) and further on create UML class diagrams from that and even create code skeletons in Python to demonstrate the full MDA cycle. They achieved best results with a finetuned BERT resulting in over 98% F1-score for class names and still very good 96% on the different kinds of relations. A more recent approach to the same issue is dubbed NOMAD and uses a multi-agent approach to generate PlantUML class diagrams from natural language requirements [12]. They also perform a detailed error analysis distinguishing missing, extra and wrong for classes and attributes. Relationship-specific errors further distinguish between extra and duplicate relations. Methods are not part of the study. They compare their multi-agent setup with the performance of the same LLM in a single-agent scenario and find the multi-agent system improves the F1-scores overall between 4% and 10% absolute. In some cases with low baseline performance, the F1-scores even got 40% better (for relationships) using GPT4o as the LLM. This is achieved mainly by eliminating duplicate, extra and missing relations. The number of wrong relationships stays relatively high compared to class and attribute errors.

### 2.2 Diagram understanding with VLMs

Different kinds of diagrams like line and bar charts pose similar challenges to VLMs for question answering. As soon as it is not sufficient to extract some labels from the diagram to answer the question correctly but visual reasoning is required, the accuracy of the models drops significantly [5]. To introduce more variety to questions, ChartQA uses real-world diagrams with human-authored questions in addition to template-based machine-generated ones. Masry et al. stress the necessity of open-ended questions that cannot be treated as a classification problem [5]. The best model only achieved 45% accuracy in their tests, whereas other contemporary datasets with similar focus were solved with 95% accuracy. However, only few years later the ChartQA benchmark was also saturated with Claude Sonnet 3.5 achieving over 90% accuracy [6]. Therefore, Masry et al. collected an even more challenging and diverse dataset from 157 different online platforms and also made sure to have more complex diagrams e.g. with stacked or grouped bars and also included some infographics. The type of questions is focused on mathematical reasoning which represents the shift in capabilities by emerging LLMs with test-time scaling in 2025. Claude Sonnet 3.5 stayed ahead of the competition in ChartQA Pro as well but reached only 55.8% accuracy using a Chain-of-thought prompting. Gemini Flash 2.0 scored 53.7% and Qwen2 VL 7B, the best open weight model, achieved 37% accuracy whereas the human baseline was 85%.

Deka and Dervereux extract JSON representations from BPMN diagrams using VLMs [8]. They compiled a dataset of 202 diagrams split into 101 for training, 50 for development and 50 for test. They achieve 80.7% F1-score with GPT 4.1 and 80.2% with Mistral 3.1 small, the best open weight model for extracting the names. With types included, F1-scores drop to 70.7% and 69.1% respectively. Similarly to the UML class diagram generation, relations were most erroneous in the experiments. GPT 4.1 scored only 47.5% here. Mistral even dropped to 39.5%. Sequence flows and message flows were among the BPMN elements with most errors.

Ranjani and Prabhudesai [7] research the correctness of VLMs translating UML sequence diagrams into PlantUML textual representations. They use Claude Sonnet 3.7 and GPT-4V and test them on 50 different diagrams with an average of 50 lines of puml script. 10 diagrams were pretty simple with under 20 lines of script as groundtruth. The authors analyze the types of errors and classify them into insertions, deletions and substitutions, split by the model elements like message, box and participant. The overall performance was rather disappointing with error rates up to 100% for box elements and 47% for participants by GPT-4. Claude Sonnet also struggled with boxes but generated only 11% errors for participants.

All in all, the examples show that even modern VLMs that perform well in many standard benchmarks, have problems with interpreting important aspects of diagrams.

# 3 Dataset

We build a dataset of UML diagrams by scraping real-world software repositories from Github. We use projects with Java as the programming language and construct UML class diagrams from them using our own converter that generates diagrams in textual PlantUML syntax which are then rendered and exported as PNG images. We filtered for projects with sufficient complexity.

**Table 1.** overview of the question types in the dataset.

| q-type | category | example question |
|---|---|---|
| Q1 | relation | Give the name of the class that lies between AbstractAppSupport and Stage in a valid relationship chain. |
| Q2 | relation | List the names of all classes that are aggregation targets of Top2Accum. Return a comma-separated list. |
| Q3 | relation | List the names of all classes that are composition targets of ClientService. Return a comma-separated list. |
| Q5 | interface | List the names of interfaces that MainStageController implements. Return a comma-separated list. |
| Q6 | attribute | List the names of private attributes declared in ClientService. Return a comma-separated list. |
| Q8 | attribute | List the names of public attributes declared in MaterialCalendarViewManager. Return a comma-separated list. |
| Q9 | method | List the names of public methods declared in CustomSplash. Return a comma-separated list. |
| Q10 | inherit | Give the name of the class that CustomSplash extends. Return only one class name. |
| Q11 | relation | List a chain of three connected classes starting from DetailsService and ending with OauthRepository. Include the relationship types between them. Format: Class A, relation1, Class B, relation2, Class C |
| Q12 | relation | Name the class with the highest number of relationships (as source or target). Write out only the class name. |
| Q14 | attribute | List the names of static attributes declared in OauthRepository. Return a comma-separated list. |
| Q15 | attribute | List the names of final attributes declared in OauthRepository. Return a comma-separated list. |
| Q16 | method | List the names of static methods declared in MainController. Return a comma-separated list. |
| Q19 | attribute | List the names of attributes in OauthRepository that are both static and final. Return a comma-separated list. |
| Q20 | relation | List the names of all classes that have no relationships of any kind (neither source nor target). Return a comma-separated list. |

| q-type | category | example question |
| --- | --- | --- |
| Q21 | relation | List the names of all classes that have a composition relationship where JdbcTemplate is the target.<br>Return a comma-separated list. |
| Q22 | relation | List the names of all classes that have an aggregation relationship where Integer is the target.<br>Return a comma-separated list. |

We compiled a list of 22 questions and collect groundtruth answers by applying them to the crawled repositories. Table 1 shows examples for questions of all 22 types.

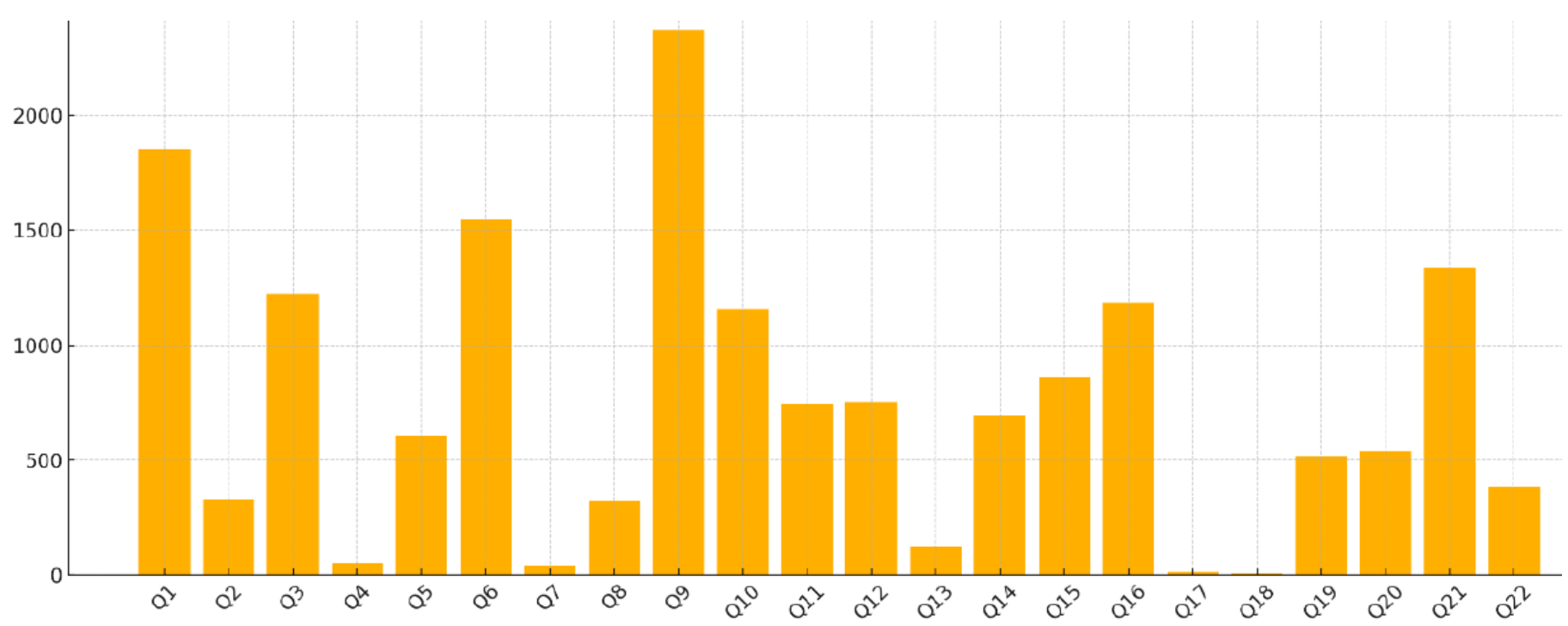

**Fig. 1.** Histogram with number of questions per question type in the original dataset.

The reason for the low number of questions for Q4, Q7, Q13, Q17 and Q18 is that the kind of relations asked for were only present in very few repositories that we crawled. To get a more balanced dataset, we removed them and additionally limited the amount of question for Q9 to 2000. The test and training dataset are published: https://huggingface.co/iisys-hof/datasets.

Image size depends on the number of classes in the diagram, as well as on the complexity of relations and to some degree on the number of attributes and methods per class. The automatic layout algorithm cannot use space as efficiently as a human could, but we did no manual optimization. We excluded images with more than 4000 pixels in any direction, which mainly affected the width as images tend to be broader than high with our settings.

## 4 Experiments

We finetune Qwen 2.5 VL 7B (baseline) on our dataset with ~16.000 question image pairs with LoRA (Low Rank Adaptation) and test on a held-out portion of the dataset of ~1600 question image pairs that also contain unseen images. We used a learning rate of 2e-4, R 96, α 96 and dropout 0.05. We compare results of 2, 4 and 8 epochs with either training only the vision tower, the LLM or both. We compare results with the baseline as well as the more recent Qwen 3 VL 8B and the larger Qwen 3.5 27B models.

We determine accuracy by a clever exact match algorithm that takes care of different sequences of answers in cases where a list of results is returned. We require correct capitalization of results since in Java a different capitalization of a class, attribute or method name means a different identifier. We list capitalization errors separately in the results analysis.

## 5 Results

The analysis shows that existing VLMs have some knowledge about UML diagrams and are able to correctly answer about half of the questions. There is moderate progress from one model generation to the next. Even the bigger Qwen 3.5 27B model scores only 65% which shows our dataset is challenging. Our best finetune with 8 epochs on both vision and LLM part of Qwen 2.5 VL 7B achieves 85.9% accuracy outperforming the much bigger Qwen 3.5 model significantly.

**Table 2:** overall results of different VLMs on our UML class diagram dataset

| Model | Qwen 2.5 VL 7B | Qwen 3 VL 8B | Qwen 3.5 27B | Best finetune |
|---|---|---|---|---|
| Result | 44.4% | 53.9% | 65.1% | 88.3% |

Analyzing the results per question, we see that only Q2 is hard to answer even for the finetune. For all other question types, the performance is at least 80%. You also see that the largest jump in accuracy is between the baseline and the 2 epoch finetune (+30%). The additional training for 4 and 8 epochs provides only smaller improvements (+7% and +5%).

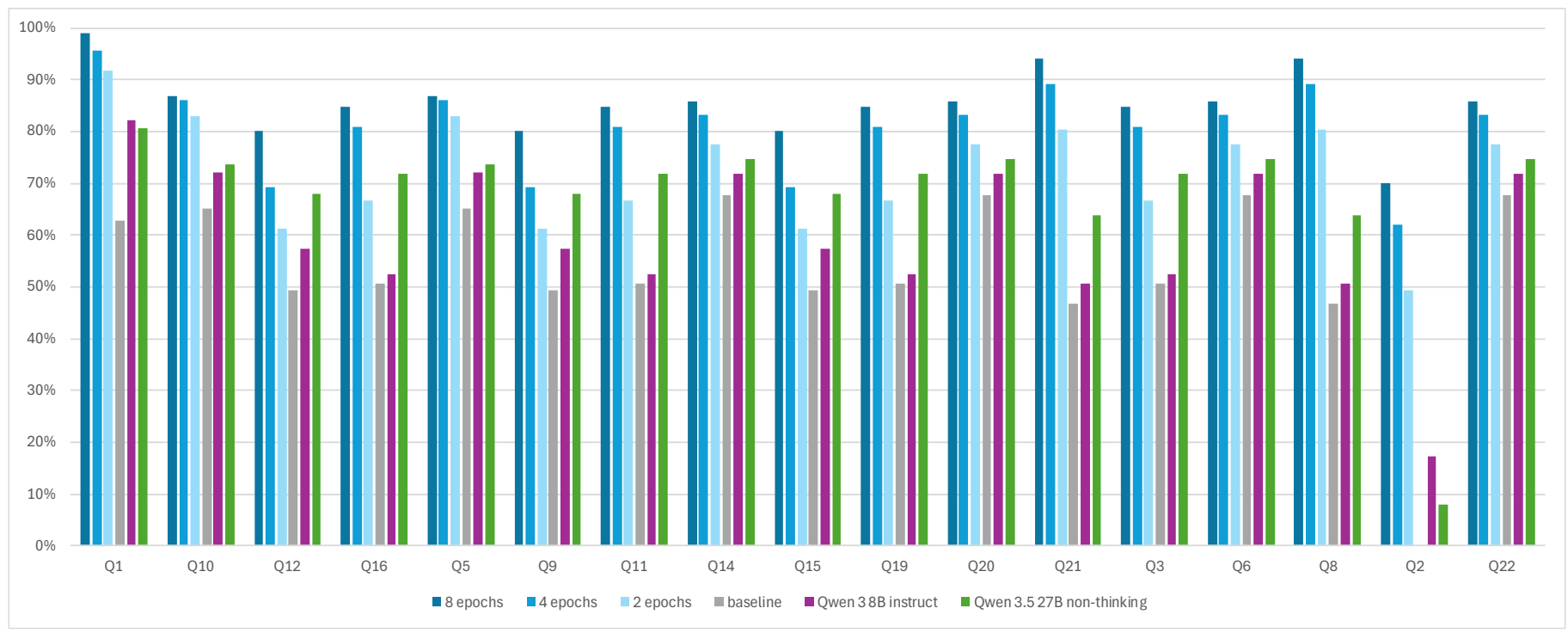

**Fig. 2.** Accuracy of our finetunes compared to reference models per question type.

We did a closer analysis of errors and also compared the vision-only and llm-only finetunes with the finetunes adjusting all parts of the model (see **Fehler! Verweisquelle konnte nicht gefunden werden.**). It is obvious that training the vision tower alone brings only minor improvements and the LLM finetuning has a much larger impact. Vision training has the largest impact on reducing the number of additional results reported. The other error categories are nearly not affected at all. Wrong capitalization is a small problem in general. Answering with an attribute or class name instead of "none" is a significant problem and mitigated only partly with the finetune (still 66 error cases out of 119). In several cases, more results than correct are returned. Missing result parts is also an issue, but less frequent. In some question types (e.g. Q11) the relation between the classes should be reported. This is problematic and frequently done wrong. The low overall percentage is misleading here because there are not too many questions where this error can occur.

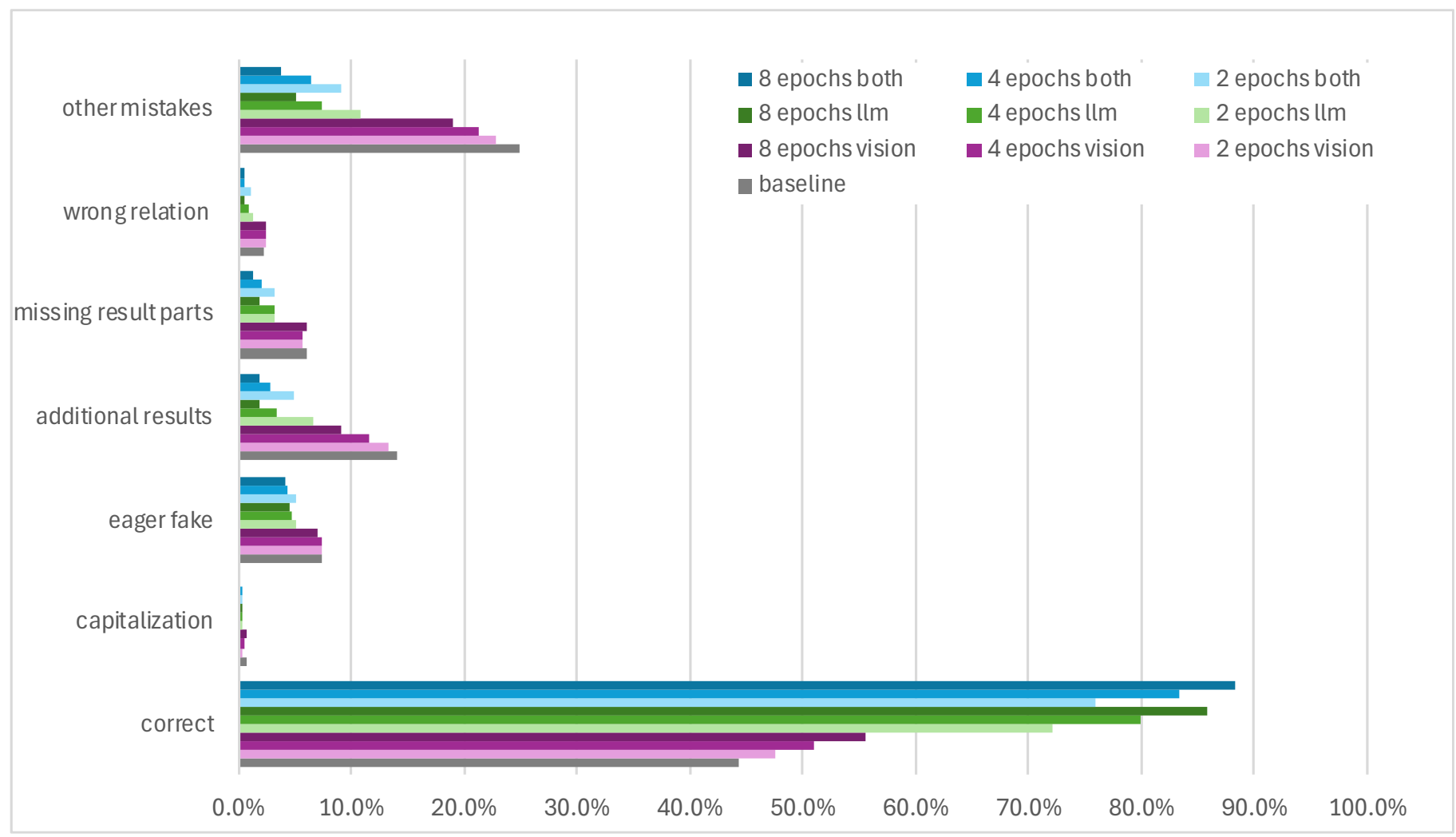

**Fig. 3.** Error analysis of different finetunes per error type.

## 6 Discussion and error analysis

**Misinterpretation of Visual Layouts**: The model occasionally groups spatially adjacent but logically unrelated classes or methods, suggesting that its positional encoding may not adequately capture spatial structure in schematic images such as UML diagrams. This is a known limitation in ViT-based encoders, which lack explicit layout modeling mechanisms [6]. Employing layout-aware encoders or spatially grounded graph representations could address this issue.

**Inability to Perform Multi-Hop Reasoning:** Errors involving reasoning chains across multiple classes (e.g., identifying an intermediate node between two indirectly connected classes) reveal the model's difficulty in maintaining coherent attention over longer token distances. Such decay in attention fidelity is consistent with observations in other VLMs applied to symbolic domains. Potential mitigation strategies include the use of chain-of-thought prompting or use of test-time scaling [18].

**Identifier Confusion Due to Naming Similarities:** The model sometimes confuses classes with similar names (e.g., User vs. UserDTO), indicating insufficient semantic disambiguation. This suggests that while the visual encoder captures token presence, it lacks grounding in language-level semantics. Integrating abstract syntax tree (AST) context or employing token-level contrastive learning may help resolve such ambiguities.

**Failure to Predict Absence of Relationships**: In cases where the correct answer is explicitly “None” (e.g., for non-existent inheritance or empty attribute lists), the model frequently outputs class names instead. This problem is well known for other vision tasks as well [19] and was one of the reasons that we included such cases. We captured that in our analysis in the category “eager fake” (see **Fig. 3**). Additional result parts are somehow related, but can be fixed by training much easier.

**Limited Multi-Object Compositional Reasoning**: When a correct answer requires aggregating information across multiple entities – such as computing the class with the highest number of relationships – the model exhibits reduced accuracy. This likely reflects a lack of internal memory or compositional reasoning capability, a common limitation of current VLMs when faced with topological or global analysis tasks.

These failure categories align with findings in recent evaluations of VLMs on structured visual data, particularly diagrams and charts, where symbolic reasoning and layout comprehension remain key challenges [17], [20].

**Capitalization** is no problem for models with only 12 error cases in the baseline (0.8%) and zero error cases remaining after 8 epochs of finetuning.

# 7 Conclusion and Future Work

Summed up, the analysis shows that even most recent models like Qwen 3.5 27B that achieves 75% in MMMU-Pro (human experts 81%) struggle with our UML benchmark. Our dataset is therefore challenging but also manageable as demonstrated by the increased accuracy of more capable models (progress from 44 to 54% and 65% from Qwen 2.5 VL 7B over Qwen3 VL 8B to Qwen 3.5 27B) and our finetune achieves 88% accuracy using a LoRA approach with only few parameters. We showed that the problems are located in the LLM part of the VLM, since finetuning the vision-tower only led to minimal increase in accuracy, whereas finetuning of the LLM part was nearly as good as finetuning the whole model (85.9% vs. 88.3%). Detecting attributes, methods or relations that fulfil a certain query when there is none present remained the most frequent error case for the finetuned model (66 cases, 4.1%) which is not even a reduction of 50% compared to the baselines 7.5% error rate. This indicates that the models still struggle to say “no” and rather tend to hallucinate answers that sound plausible. This is in line with previous results [19] and might also be attributed to the existing benchmarks that do not include adversarial examples to measure exactly that kind of hallucinations. Future research should systematically develop similar large scale training datasets for other well-known diagram types like BPMN, UML use case, or ARIS event-driven process chains to build a better general understanding for all kinds of diagrams. We hypothesize that there should be positive spill-over effects between those use cases.

**Acknowledgments.** This work was partly funded by the European Fonds for Regional Development (EFRD) in the M4-SKI project (multi-modal man-machine interface with AI).

## References

[1] A. Radford *et al.*, “Learning transferable visual models from natural language supervision,” *International Conference on Machine Learning*, pp. 8748–8763, 2021.

[2] M. Tschannen *et al.*, “SigLIP 2: Multilingual Vision-Language Encoders with Improved Semantic Understanding, Localization, and Dense Features,” Feb. 20, 2025, *arXiv*: arXiv:2502.14786. doi: 10.48550/arXiv.2502.14786.

[3] M. Caron *et al.*, “Emerging properties in self-supervised vision transformers,” in *Proceedings of the IEEE/CVF international conference on computer vision*, 2021, pp. 9650–9660. Accessed: Jan. 26, 2026. [Online]. Available: https://openaccess.thecvf.com/content/ICCV2021/html/Caron_Emerging_Properties_in_Self-Supervised_Vision_Transformers_ICCV_2021_paper

[4] M. Hinck, M. L. Olson, D. Cobbley, S.-Y. Tseng, and V. Lal, “LLaVA-Gemma: Accelerating Multimodal Foundation Models with a Compact Language Model,” Jun. 10, 2024, *arXiv*: arXiv:2404.01331. doi: 10.48550/arXiv.2404.01331.

[5] A. Masry, X. L. Do, J. Q. Tan, S. Joty, and E. Hoque, “Chartqa: A benchmark for question answering about charts with visual and logical reasoning,” in *Findings of the association for computational linguistics: ACL 2022*, 2022, pp. 2263–2279. Accessed: Jan. 26, 2026. [Online]. Available: https://aclanthology.org/2022.findings-acl.177/

[6] A. Masry *et al.*, “ChartQAPro: A More Diverse and Challenging Benchmark for Chart Question Answering,” Apr. 10, 2025, *arXiv*: arXiv:2504.05506. doi: 10.48550/arXiv.2504.05506.

[7] H. G. Ranjani and R. Prabhudesai, “Measuring Visual Understanding in Telecom domain: Performance Metrics for Image-to-UML conversion using VLMs,” in *Proceedings of the 5th Workshop on Evaluation and Comparison of NLP Systems*, 2025, pp. 9–20. Accessed: Jan. 26, 2026. [Online]. Available: https://aclanthology.org/2025.eval4nlp-1.2/

[8] P. Deka and B. Devereux, “Structured Extraction from Business Process Diagrams Using Vision-Language Models,” Nov. 27, 2025, *arXiv*: arXiv:2511.22448. doi: 10.48550/arXiv.2511.22448.

[9] Z. Babaalla, A. Jakimi, and M. Oualla, “LLM-Driven MDA Pipeline for Generating UML Class Diagrams and Code,” *IEEE Access*, 2025, Accessed: Jan. 26, 2026. [Online]. Available: https://ieeexplore.ieee.org/abstract/document/11184520/

[10] A. Conrardy and J. Cabot, “From Image to UML: First Results of Image Based UML Diagram Generation Using LLMs,” Apr. 17, 2024, *arXiv*: arXiv:2404.11376. Accessed: Apr. 28, 2024. [Online]. Available: http://arxiv.org/abs/2404.11376

[11] D. De Bari, “Evaluating Large Language Models in Software Design: A Comparative Analysis of UML Class Diagram Generation,” PhD Thesis, Politecnico di Torino, 2024. Accessed: Apr. 28, 2024. [Online]. Available: https://webthesis.biblio.polito.it/31177/

[12] P. Giannouris and S. Ananiadou, “NOMAD: A Multi-Agent LLM System for UML Class Diagram Generation from Natural Language Requirements,” Nov. 27, 2025, *arXiv*: arXiv:2511.22409. doi: 10.48550/arXiv.2511.22409.

[13] V.-V. Nguyen, H.-K. Nguyen, K.-S. Nguyen, M.-H. Luong Thi, T.-V. Nguyen, and D.-Q. Vu, “Automated UML Generation: A Framework for Class Diagram Synthesis and Multimodal Validation,” in *Future Data and Security Engineering*, vol. 2709, T. K. Dang, J. Küng, and T. M. Chung, Eds., in Communications in Computer and Information Science, vol. 2709. , Singapore: Springer Nature Singapore, 2026, pp. 212–224. doi: 10.1007/978-981-95-4724-1_15.

[14] V.-V. Nguyen, H.-K. Nguyen, K.-S. Nguyen, H. Luong Thi Minh, T.-V. Nguyen, and D.-Q. Vu, “A Novel Pipeline for Automatic UML Sequence Diagram Synthesis and Multimodal Scoring,” in *Intelligent Systems and Data Science*, vol. 2713, N. Thai-Nghe, T.-N. Do, and S. Benferhat, Eds., in Communications in Computer and Information Science, vol. 2713. , Singapore: Springer Nature Singapore, 2026, pp. 473–485. doi: 10.1007/978-981-95-3355-8_34.

[15] R. Soldati, “LLM-based Generation and Evaluation of UML Class Diagrams,” PhD Thesis, Politecnico di Torino, 2025. Accessed: Jan. 26, 2026. [Online]. Available: https://webthesis.biblio.polito.it/35962/

[16] E. J. Hu *et al.*, “Lora: Low-rank adaptation of large language models.,” *ICLR*, vol. 1, no. 2, p. 3, 2022.

[17] S. Mukhopadhyay, A. Qidwai, A. Garimella, P. Ramu, V. Gupta, and D. Roth, “Unraveling the truth: Do VLMs really understand charts? a deep dive into consistency and robustness,” in *Findings of the Association for Computational Linguistics: EMNLP 2024*, 2024, pp. 16696–16717. Accessed: Jan. 26, 2026. [Online]. Available: https://aclanthology.org/2024.findings-emnlp.973/

[18] M. Ahmadpour, A. Meighani, P. Taebi, O. Ghahroodi, A. Izadi, and M. S. Baghshah, “Limits and Gains of Test-Time Scaling in Vision-Language Reasoning,” Dec. 11, 2025, *arXiv*: arXiv:2512.11109. doi: 10.48550/arXiv.2512.11109.

[19] R. Peinl and V. Tischler, “VLM@school: Evaluation of AI Image Understanding on German Middle School Knowledge,” in *Proceedings of the Future Technologies Conference (FTC) 2025, Volume 1*, vol. 1675, K. Arai, Ed., in Lecture Notes in Networks and Systems, vol. 1675. , Cham: Springer Nature Switzerland, 2025, pp. 664–680. doi: 10.1007/978-3-032-07986-2_41.

[20] S. Joshi *et al.*, “MM-GEN: Enhancing Task Performance Through Targeted Multimodal Data Curation,” Jan. 07, 2025, *arXiv*: arXiv:2501.04155. doi: 10.48550/arXiv.2501.04155.